\documentclass[letterpaper]{article} 
\usepackage{aaai2027}  
\usepackage[hyphens]{url}  
\usepackage{graphicx} 
\usepackage{natbib}  
\usepackage{caption} 
\usepackage{algorithm}
\usepackage{algorithmic}
\usepackage{amsmath}
\usepackage{amssymb}
\usepackage{booktabs}
\usepackage{multirow}
\usepackage{pifont}

\newcommand{\bl}{\mathit{bl}}
\newcommand{\flowblock}{FlowBlock}
\newcommand{\masktok}{\texttt{[M]}}
\newcommand{\thspawn}{\theta_{\mathrm{spawn}}}
\newcommand{\taumask}{\tau_{\mathrm{mask}}}
\newcommand{\tauedit}{\tau_{\mathrm{edit}}}

\title{FlowBlock: Wavefront-Parallel Decoding for Self-Correcting Diffusion Language Models}

\author{
    Bing Tian\textsuperscript{$\dag$}\quad
    Haikun Liu\textsuperscript{$\dag$}\quad
    Xiaocheng Zhong\textsuperscript{$\S$}\quad
    Zhuohui Duan\textsuperscript{$\dag$}\quad
    Zhaokai Luo\textsuperscript{$\S$}\quad
    Huayi Jin\textsuperscript{$\S$}\quad
    Zhiyong Wang\textsuperscript{$\S$}\quad
    Xiaofei Liao\textsuperscript{$\dag$}
}

\affiliations{
    \textsuperscript{$\dag$}\textit{
        Huazhong University of Science and Technology
    }
    \qquad
    \textsuperscript{$\S$}\textit{
        Xiaohongshu Inc., China
    }
}

\begin{document}

\maketitle

\begin{abstract}
Block-wise diffusion large language models (dLLMs) decode sequentially at the block level, enabling effective KV-cache reuse across blocks but making inter-block decoding strictly serial. Prior work has attempted to unlock inter-block parallelism through post-training methods, but achieves only modest speedups and often degrades accuracy. We observe that self-correcting dLLMs offer a training-free alternative: token-to-token (T2T) editing can repair tokens drafted with a slightly stale upstream context, so a downstream block requires only an informative draft rather than a finalized predecessor. This turns block finality from a hard dependency into a scheduling resource. We propose \textbf{\flowblock{}}, a training-free parallel decoding framework built on two mechanisms. (i) \emph{Gated Wavefront Decoding} admits blocks into a bounded wavefront only when a readiness gate is satisfied, jointly refines active blocks via T2T editing, and commits blocks in order under a windowed block-causal mask that preserves exact frozen-prefix KV caches reuse. (ii) \emph{Heterogeneous Wavefront Packing} assigns each request an independent wavefront while packing asynchronous windows into dense, shape-stable batched forwards. Across different benchmarks, \flowblock{} improves tokens per second (TPS) over LLaDA-2.1 and LLaDA-2.0, two serial block-wise dLLMs, by up to 2.95$\times$ and 4.01$\times$, while reducing latency by up to 53.6\% and 77.1\%, respectively. It also improves average accuracy by 1.3 points. Compared with D2F, a training-based inter-block-parallel baseline, \flowblock{} achieves higher accuracy and up to 16$\times$ higher batched serving throughput.
\end{abstract}

\begin{links}
    \link{Code}{https://github.com/Red-EAD/FlowBlock}
\end{links}

\section{Introduction}
\label{sec:intro}

Diffusion large language models (dLLMs) generate text by iteratively  denoising multiple positions per forward pass \citep{austin2021d3pm,li2022diffusionlm,lou2024sedd,sahoo2024mdlm,nie2025llada,ye2025dream}. This bidirectional denoising paradigm reduces the strict token-by-token dependency of auto-regressive decoding, but limiting exact KV-cache reuse due to every position's KV cache state is refreshed at each step~\citep{wu2025fastdllm,ma2025dkvcache}. Block-wise dLLMs \citep{han2023ssdlm,arriola2025block,inclusionai2025llada2} address   this issue by decoding fixed-length blocks from left to right: the active block is denoised in parallel, and committed blocks are frozen as an AR-style prefix cache. Recently, self-correcting dLLMs, such as LLaDA-2.1 \citep{inclusionai2026llada21}, further proposes token-to-token (T2T) editing, which revises already revealed tokens and substantially increases intra-block parallelism.

Despite these advances, block-wise dLLMs remain bottlenecked by inter-block serialization. A downstream block $B_{g+1}$ cannot begin until its predecessor $B_g$ has been fully finalized, leaving the late denoising steps of $B_g$ unable to advance subsequent blocks. D2F explores inter-block parallelism through post-training by distilling a checkpoint capable of denoising downstream blocks from incomplete upstream context, but its speedup remains limited and comes at the cost of degraded accuracy. We ask whether inter-block parallelism can be exposed at inference time on the unmodified self-correcting dLLMs. Our key observation is that T2T editing weakens the dependency between adjacent blocks: a downstream block only needs an informative upstream draft, not a finalized predecessor, because later edits can repair tokens produced under slightly stale context. Therefore, block finality can be treated as a schedulable resource rather than a hard execution precondition.

However, turning this relaxation into practical speedup is nontrivial. \emph{First, overlap must be safe and cache-correct.} Premature overlap may create low-quality drafts and erase any speedup through extra correction cost, while an incorrect attention pattern would invalidate frozen KV-cache reuse. \emph{Second, wavefront execution must remain efficient under batched serving.} Different requests become ready and finish at different steps, so a batch-synchronous wavefront stalls on stragglers, whereas fully independent execution produces irregular tensor shapes that GPUs work inefficiently.

In this paper, we present \textbf{\flowblock{}}, a training-free execution framework that converts block-serial decoding process of self-correcting dLLMs into wavefront-parallel decoding. \flowblock{} built on two techniques that address the challenges above. \textbf{Gated Wavefront Decoding} jointly refines a bounded window of live blocks, admits a new block only when the frontier readiness exceeds $\thspawn$, and commits blocks in order under a windowed block-causal mask that preserves exact frozen-prefix KV-cache reuse. \textbf{Heterogeneous Wavefront Packing} gives each request its own wavefront state while packing all active windows into dense $[B,q]$ forwards using absolute positions and per-row block-diagonal masks. Together, \flowblock{} exposes inter-block parallelism without retraining while retaining the regularity required for efficient batched serving.

We implement \flowblock{} in dInfer \citep{inclusionai2025dinfer}, which is an SGLang-backed inference framework for dLLMs, and evaluate it with LLaDA-2.1-mini on different benchmarks. At batch size 1, \flowblock{} improves TPS by up to $1.57\times$ and $3.17\times$ over LLaDA-2.1 and LLaDA-2.0, respectively. Under batched serving, the gains increase to $2.95\times$ and $4.01\times$, with latency reduced by up to $53.6\%$ and $77.1\%$. Compared with D2F, \flowblock{} is both more accurate and more scalable, reaching up to $16\times$ higher batched throughput. Across all benchmarks it matches or improves baseline accuracy, raising the average by $1.3$ points.

In summary, our major contributions are as follows.
\begin{itemize}
\item We identify a structural opportunity in self-correcting dLLMs: T2T editing relaxes inter-block dependencies, enabling training-free scheduling during inference.
\item We propose Gated Wavefront Decoding, which combines readiness-gated admission, joint T2T refinement, and a windowed block-causal mask to overlap blocks while preserving exact frozen-prefix KV reuse.
\item We propose Heterogeneous Wavefront Packing, which turns asynchronous per-request wavefronts into dense, shape-stable batched forwards through per-row gathers, absolute-position KV indexing, and block-diagonal attention.
\item We implement \flowblock{} and demonstrate consistent throughput and latency improvements at matched-or-better accuracy across different benchmarks.
\end{itemize}

\section{Related Work}
\label{sec:related}

\paragraph{Diffusion language models.}
Discrete diffusion language models~\citep{austin2021d3pm,li2022diffusionlm,lou2024sedd,sahoo2024mdlm} formulate token generation as iterative denoising from masked sequences, enabling multiple positions to be decoded in parallel. Recent models such as LLaDA \citep{nie2025llada}, Dream \citep{ye2025dream}, Mercury \citep{inceptionlabs2025mercury}, and Seed Diffusion \citep{bytedance2025seeddiffusion} show that this paradigm can approach auto-regressive model quality while offering substantial decoding speedups. However, vanilla dLLMs typically perform bidirectional full-sequence denoising, where all positions refresh their KV-cache at every step. This prevents exact KV-cache reuse and incurs $O(L^2)$ attention per step. 

\paragraph{Block-wise diffusion LLMs.}
Block-wise dLLMs~\citep{han2023ssdlm,arriola2025block} recover exact prefix-cache reuse by partitioning the generation region into fixed-length blocks and orderly decoding them. Positions within the active block are denoised in parallel, while the KV entries of committed blocks are frozen and reused as an autoregressive-style prefix cache. LLaDA-2.0 \citep{inclusionai2025llada2} scales this recipe, and LLaDA-2.1 \citep{inclusionai2026llada21} further improves the paradigm with token-to-token (T2T) editing for intra-block self-correction. However, these methods still serialize decoding across blocks. D2F \citep{wang2025d2f} exposes inter-block parallelism through post-training, distilling a model to denoise downstream blocks from incomplete upstream context. In contrast, \flowblock{} exploits the self-correcting behavior already present in T2T-enabled dLLMs. It pipelines blocks at inference time by admitting only readiness-qualified downstream drafts, preserves exact KV reuse for committed prefixes, and packs asynchronously progressing requests into dense batched forward passes. As shown in \S\ref{sec:exp}, this training-free design improves accuracy and scales more effectively under batched serving than the training-based D2F baseline.

\section{Background and Motivation}
\label{sec:background}

\subsection{Block-Wise Decoding with Self-Correction}

\paragraph{Joint M2T--T2T decoding.}
Conventional absorbing-state dLLMs make each mask-to-token transition irreversible: once a token is revealed, later context cannot directly revise it. Self-Correcting dLLMs (LLaDA-2.1 series) instead trains an editable ``Draft-and-Edit'' decoder that supports both token generation and retrospective correction~\citep{inclusionai2026llada21}. During continual pre-training and supervised fine-tuning, it uses a mixture of M2T and T2T objectives. The M2T stream learns to recover masked tokens, whereas the T2T stream perturbs already observed tokens with stochastic noise and trains the model to restore the clean sequence. Consequently, a single checkpoint learns to act as both a fast drafter and an editor. \flowblock{} relies on this pretrained editability and requires no additional model training.

Let $x_t$ denote the current sequence state at denoising step $t$, and let $\mathcal{P}$ be the set of fixed prompt positions. For each position $i$, let $\hat{x}_{0,i}$ denote the top-1 prediction under the current logits and let $p_i$ be its softmax confidence. LLaDA-2.1 identifies two update sets:
\begin{equation}
\label{eq:joint}
\begin{aligned}
\Gamma_t &= \{\, i \;:\; x_t[i] = \masktok
                 \,\wedge\, p_i > \taumask \,\}, \\
\Delta_t &= \{\, i \;:\; x_t[i] \neq \hat{x}_{0,i}
                 \,\wedge\, p_i > \tauedit
                 \,\wedge\, i \notin \mathcal{P} \,\}.
\end{aligned}
\end{equation}
The next state sets $x_{t+1}[i]=\hat{x}_{0,i}$ for
$i\in\Gamma_t\cup\Delta_t$ and leaves all other positions unchanged.
The update type is determined by the current state: updating a masked
position performs mask-to-token (M2T) drafting, whereas updating an
already revealed non-prompt position performs token-to-token (T2T)
editing. Unlike remasking-based correction, T2T directly replaces the
current token when a different prediction becomes sufficiently
confident.

The two thresholds decouple drafting from correction. The M2T threshold
$\taumask$ controls how aggressively new tokens are revealed, while
$\tauedit$ controls when an existing token may be overwritten. Lower
drafting thresholds expose more positions per forward pass and rely on
later T2T updates to repair premature decisions; more conservative
thresholds favor output stability. Because every denoising step
recomputes logits under the latest block state, newly revealed context
can change the preferred token at an earlier position and trigger a
subsequent edit. Once no masks remain, decoding continues for up to
$S_{\mathrm{pe}}$ post-edit steps and may stop early when a step produces
no update. The resulting locally finished state serves as the
block-level commit signal.

\begin{figure*}[t]
\centering
\includegraphics[width=0.99\textwidth]{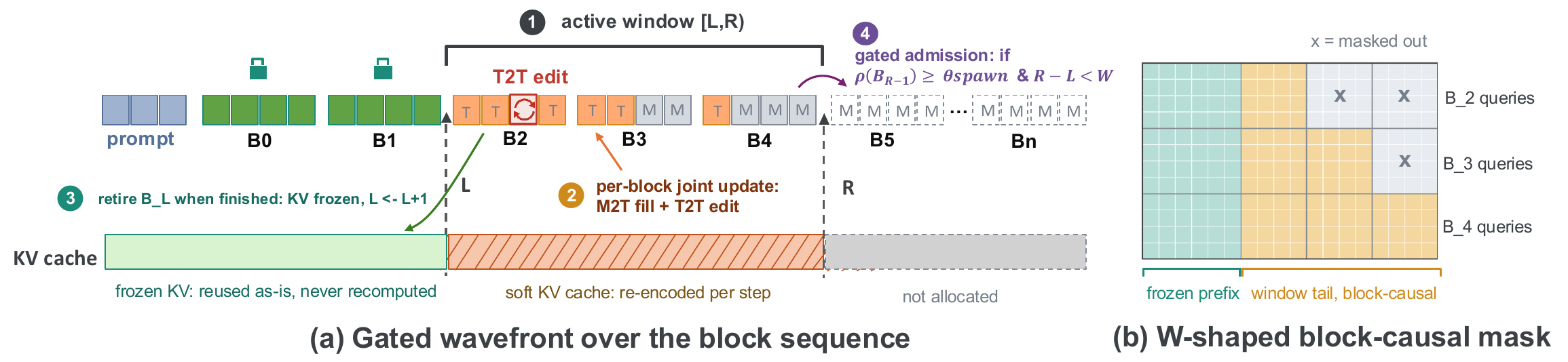}
\caption{Gated Wavefront Decoding. (a) A sliding active window $[L,R)$ holds up to $W$ adjacent blocks at different denoising stages over a tail-aligned cache. Each step: \ding{182} one forward decodes the whole window; \ding{183} every active block applies the joint M2T\,$\cup$\,T2T update of Eq.~\eqref{eq:joint}; \ding{184} the leftmost block retires once locally finished, freezing its KV into the committed prefix; \ding{185} the gate admits the next block only when the frontier readiness exceeds $\thspawn$ (Eq.~\eqref{eq:admit}). KV caches are mutable \emph{only inside} the window, and all committed prefixes are frozen and reused exactly. (b) The W-shaped block-causal mask: every window query sees the full committed prefix and earlier in-window blocks, but never future blocks.}
\label{fig:overview}
\end{figure*}


\subsection{Motivation: Self-Correction Unlocks Inter-Block Parallelism}
\label{sec:motivation}

The self-correction mechanism above improves parallelism \emph{within} a block, but standard block-wise decoding still enforces a strict dependency \emph{between} blocks. Block $B_{g+1}$ can start only after $B_g$ is locally finished and committed. This schedule is safe but conservative: the late denoising steps of $B_g$ often refine only a small number of uncertain positions, while all downstream blocks remain idle.

The key opportunity is that T2T editing makes early drafts mutable rather than irrevocable. A downstream block generated under a slightly stale upstream draft can still be revised in later denoising steps as the upstream block becomes more stable. Thus, a downstream block does not always require a finalized upstream block; it only requires an upstream draft that is informative enough to support reliable refinement. This turns inter-block finality from a hard data dependency into a scheduling decision.

\begin{figure*}[t]
\centering
\includegraphics[width=0.99\textwidth]{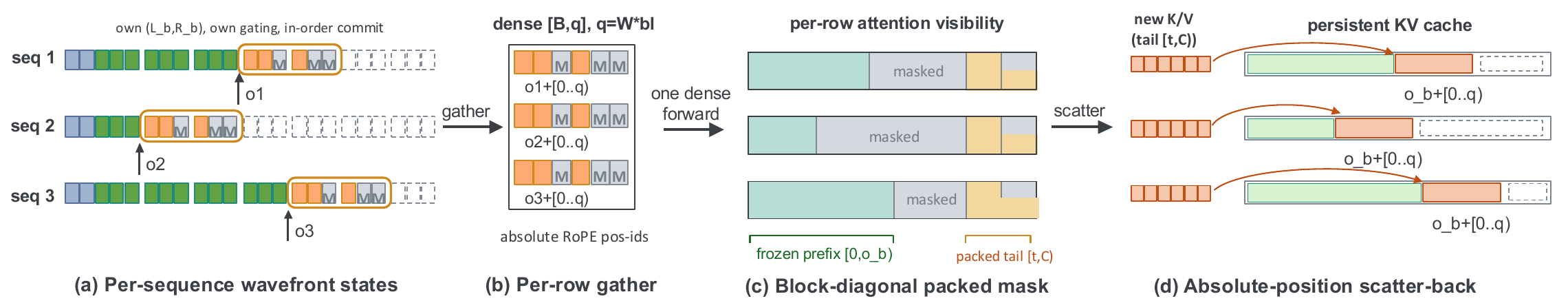}
\caption{Heterogeneous Wavefront Packing. (a) Each sequence advances its own wavefront $(L_b, R_b)$ at its own rhythm. (b) Because every window has fixed capacity $q = W\!\cdot\!\bl$, a per-row gather at offsets $o_b$ yields a dense $[B,q]$ batch---shape-stable, no padding---with absolute RoPE position ids. (c) A per-row block-diagonal mask routes each row to its own frozen prefix $[0,o_b)$ and its own block-causal tail, masking the stale middle; batched attention is row-independent, so cross-sequence isolation is structural. (d) New window KV is scattered back to per-row absolute offsets of one preallocated cache.}
\label{fig:hwp}
\end{figure*}

\section{FlowBlock Framework}
\label{sec:method}

\flowblock{} reinterprets block-wise decoding as a \emph{self-timed dataflow}: blocks act as pipeline stages, the active window forms a wavefront that advances from left to right, and \emph{admission} and \emph{retirement} are triggered by the decoding state rather than a fixed schedule. \flowblock{} operates purely at the execution level: it requires no post-training, no kernels modifications, and relies on the model's intrinsic T2T self-correction to maintain cross-block consistency. Its two components address the challenges identified in \S\ref{sec:intro}: Gated Wavefront Decoding (\S\ref{sec:gwd}, Figure~\ref{fig:overview}) enables \emph{safe, KV-cache-correct} inter-block overlap, while Heterogeneous Wavefront Packing (\S\ref{sec:hwp}, Figure~\ref{fig:hwp}) translates this overlap into \emph{efficient batched serving}.

\subsection{Gated Wavefront Decoding}
\label{sec:gwd}

\paragraph{Wavefront window.}
GWD replaces the single active block with a window $[L,R)$ of at most $W$ adjacent blocks that denoise concurrently (Figure~\ref{fig:overview}a). Each step performs one forward over all $W\!\cdot\!\bl$ window positions, tail-aligned after the frozen prefix, and each in-window block applies the joint M2T\,$\cup$\,T2T update of Eq.~\eqref{eq:joint} to its own logits. An active upstream block is exposed to its downstream blocks through its current draft: revealed tokens are fed back as ordinary embeddings and remain editable by subsequent T2T updates. Consequently, no auxiliary draft representation or model adaptation is required. Inter-block overlap is enabled entirely by the model's native self-correction rather than post-training or distillation.

\paragraph{Readiness-gated admission.}
The key question is \emph{when} to admit the next block $B_R$. Admitting too early produces unreliable drafts and excessive correction, whereas admitting too late reduces the wavefront to serial decoding. GWD measures the readiness of the frontier block $B_{R-1}$ by
\begin{equation}
\label{eq:rho}
\rho(B) \;=\; \frac{\bigl|\{\, i \in \mathcal{M}_B : p_i \ge \taumask \,\}\bigr|}{\bigl|\mathcal{M}_B\bigr|},
\end{equation}
the fraction of still-masked positions $\mathcal{M}_B$ already decodable at the model threshold ($\rho \triangleq 1$ when $\mathcal{M}_B = \emptyset$), and admits $B_R$ if
\begin{equation}
\label{eq:admit}
\rho(B_{R-1}) \ge \thspawn \;\wedge\; (R - L < W) ,
\end{equation}
The gate controls only \emph{when} a block joins the wavefront. Once admitted, every block follows the same M2T,$\cup$,T2T update rule, while remaining inconsistencies are corrected through subsequent T2T editing. The threshold $\thspawn$ therefore provides a simple speed--accuracy trade-off (\S\ref{sec:sens}), and incurs no additional computation because $\rho$ is computed directly from the current-step logits.

\paragraph{W-shaped block-causal attention.}
KV-cache correctness is enforced by a windowed block-causal mask of shape $[\,W\!\cdot\!\bl,\, C\,]$ (Figure~\ref{fig:overview}b). Queries in window block $g$ attend to the committed prefix $[0,o_L)$ and blocks $L,\dots,g$, but never to future blocks, while remaining bidirectional within the same block as in serial decoding. Since no query depends on blocks to its right, later decoding cannot invalidate the KV states of committed blocks. Mutable context is therefore confined to the live window, preserving the global frozen-prefix cache invariant.

\paragraph{In-order retirement and exact KV reuse.}
The leftmost block retires immediately after it is locally finished: its tokens become final, its KV states are frozen into the committed prefix, and $L$ advances. Because retirement occurs only when the block occupies the leftmost window position, its attention visibility exactly matches that of serial block-wise decoding. The resulting KV cache is therefore bit-identical to that obtained by re-encoding the finalized block, eliminating any refresh pass. Unlike approximate-cache methods, \flowblock{} preserves exact frozen-prefix KV reuse while confining approximation to the transient wavefront. Generation terminates once a committed end-of-sequence token is produced. Algorithm~\ref{alg:gwd} summarizes the decoding procedure.

\begin{algorithm}[t]
\caption{Gated Wavefront Decoding (one sequence)}
\label{alg:gwd}
\begin{algorithmic}[1]
\REQUIRE prompt $c$, blocks $B_0{,}\dots{,}B_{n-1}$, width $W$, gate $\thspawn$
\STATE prefill $c$; $L \leftarrow 0$; $R \leftarrow 1$ \COMMENT{$B_0$ admitted unconditionally}
\WHILE{$L < n$ \AND no EOS committed}
  \STATE $Z \leftarrow f_\theta\bigl(x_{[o_L,\, o_{R-1}+\bl)};\ \text{pos};\ \text{mask};\ \text{KV}\bigr)$ \COMMENT{one W-shaped forward}
  \FOR{$g = L$ \TO $R-1$}
    \STATE apply M2T\,$\cup$\,T2T update (Eq.~\eqref{eq:joint}) to $B_g$ from $Z_g$
  \ENDFOR
  \WHILE{$L < R$ \AND $B_L$ locally finished}
    \STATE freeze KV of $B_L$ into prefix; $L \leftarrow L + 1$ \COMMENT{retire in order}
  \ENDWHILE
  \WHILE{$R-L < W$ \AND $R < n$ \AND $\rho(B_{R-1}) \ge \thspawn$}
    \STATE $R \leftarrow R + 1$ \COMMENT{gated admission, Eq.~\eqref{eq:admit}}
  \ENDWHILE
\ENDWHILE
\end{algorithmic}
\end{algorithm}

\subsection{Heterogeneous Wavefront Packing}
\label{sec:hwp}

\paragraph{From asynchronous states to dense tensors.}
GWD naturally progresses asynchronously: readiness and completion depend on each sequence's decoding state. A batch-synchronous implementation that shares a single window $[L,R)$ must retire blocks by \texttt{finished.all()} and admit ones according to $\min_b \rho_b$, causing the entire batch to stall on the slowest request (Figure~\ref{fig:hwpabl}). HWP removes this coupling by maintaining independent wavefront states $(L_b,R_b)$, finished flags, and in-order commit pointer for each sequence(Figure~\ref{fig:hwp}a). Although blocks within different sequences progress independently, each sequence still commits blocks strictly in order.

The key packing invariant is that every window has the same max capacity $q = W\!\cdot\!\bl$; only its absolute offset $o_b = \mathrm{start}(B_{L_b})$ differs. HWP therefore gathers (Figure~\ref{fig:hwp}b)
\begin{equation}
\label{eq:gather}
\begin{aligned}
\texttt{qtok}[b] &= x\bigl[b,\ o_b + [0..q)\bigr], \quad
\texttt{pos}[b] = o_b + [0..q),
\end{aligned}
\end{equation}
into a dense, padding-free $[B,q]$ tensor while preserving absolute position ids for RoPE. As a result, the entire barch is processed by a single embedding lookup and forward pass, and the fixed tensor shape avoids recompilation and allocator churn.

\paragraph{Block-diagonal packed attention.}
Packing changes only the attention mask. Let $t=C-q$ denote the start of the shared physical tail and $s=\lfloor i/\bl \rfloor$ the window slot of query $i$. The per-row mask is defined as
\begin{equation}
\label{eq:mask}
M[b,i,j] \;=\; \underbrace{\mathbf{1}\bigl[\, j < o_b \,\bigr]}_{\text{own frozen prefix}} \;\vee\; \underbrace{\mathbf{1}\bigl[\, t \le j < t + (s{+}1)\,\bl \,\bigr]}_{\text{own window tail, block-causal}} ,
\end{equation}
which allows each query to attend only to its own committed prefix and the block-causal portion of its packed wavefront, while masking both the stale middle region $[o_b,t)$ and future blocks (Figure~\ref{fig:hwp}c). Cross-sequence isolation follows naturally because attention is computed independently for each batch row. Although all wavefronts share the same physical tail, RoPE is applied with absolute position indices, and the mask maps each row to its own logical prefix and wavefront. As a result, packed execution is token-exact with independent per-sequence execution.

\begin{table*}[t]
\centering
\small
\setlength{\tabcolsep}{2.6pt}
\begin{tabular*}{\textwidth}{@{\extracolsep{\fill}}l cccc cccc cccc cccc}
\toprule
& \multicolumn{4}{c}{LLaDA-2.0} & \multicolumn{4}{c}{LLaDA-2.1} & \multicolumn{4}{c}{D2F} & \multicolumn{4}{c}{\textbf{\flowblock{}} } \\
\cmidrule(lr){2-5} \cmidrule(lr){6-9} \cmidrule(lr){10-13} \cmidrule(lr){14-17}
Benchmark & Acc & TPF & TPS & Lat & Acc & TPF & TPS & Lat & Acc & TPF & TPS & Lat & Acc & TPF & TPS & Lat \\
\midrule
GSM8K            & 92.49 & 2.04 & 82.6  & 4.17  & 92.49 & 4.46 & 176.6 & 2.05 & 85.82 & 2.69 & 98.9  & 3.08 & \textbf{92.65} & \textbf{6.73}  & \textbf{254.2} & \textbf{1.34} \\
MATH500          & 73.40 & 2.59 & 104.6 & 11.25 & \textbf{77.00} & 5.34 & 211.7 & 5.40 & 61.20 & 2.97 & 105.0 & 9.67 & 76.20 & \textbf{8.86}  & \textbf{332.1} & \textbf{3.44} \\
Minerva-Algebra  & 91.24 & 3.00 & 120.9 & 6.50  & \textbf{93.60} & 6.29 & 247.6 & 2.87 & 81.30 & 3.55 & 127.9 & 5.20 & 93.51 & \textbf{10.14} & \textbf{378.9} & \textbf{1.81} \\
ASDIV            & 92.36 & 2.03 & 81.9  & 2.94  & \textbf{92.89} & 4.43 & 174.5 & 1.50 & 75.31 & 2.75 & 100.3 & 1.74 & 92.80 & \textbf{6.36}  & \textbf{239.7} & \textbf{1.01} \\
\midrule
HumanEval        & 84.76 & 4.37 & 159.6 & 2.16  & 82.93 & 4.93 & 180.1 & 1.79 & 66.46 & 3.20 & 112.7 & 4.80 & \textbf{86.59} & \textbf{8.41}  & \textbf{279.4} & \textbf{1.15} \\
HumanEval+       & 79.27 & 4.37 & 162.5 & 2.13  & 77.44 & 4.93 & 182.8 & 1.76 & 60.37 & 3.17 & 105.2 & 5.15 & \textbf{82.32} & \textbf{8.41}  & \textbf{282.4} & \textbf{1.14} \\
MBPP             & 79.16 & 2.75 & 105.1 & 2.24  & 82.20 & 3.26 & 124.2 & 2.09 & 68.62 & 3.40 & 122.5 & 5.33 & \textbf{83.61} & \textbf{4.43}  & \textbf{162.1} & \textbf{1.53} \\
MBPP+            & 84.39 & 2.78 & 108.7 & 2.18  & 87.04 & 3.30 & 125.8 & 1.96 & 65.08 & 3.39 & 118.4 & 4.37 & \textbf{88.36} & \textbf{4.47}  & \textbf{161.3} & \textbf{1.39} \\
\midrule
Average          & 84.63 & 2.99 & 115.7 & 4.20  & 85.70 & 4.62 & 177.9 & 2.43 & 70.52 & 3.14 & 111.4 & 4.92 & \textbf{87.00} & \textbf{7.23}  & \textbf{261.2} & \textbf{1.60} \\
\bottomrule
\end{tabular*}
\caption{Main results at batch size 1 (block length 32, generation length 2048). }
\label{tab:main}
\end{table*}

\paragraph{KV retrieval and write-back.}
KV caches are preallocated at maximum sequence length, with each row storing KV at its absolute positions. Retrieval requires no data movement: every row reads the same physical prefix slice, while the attention mask determines which positions are visible. After each forward pass, fresh window KV states are scattered from the shared tail back to their absolute locations (Figure~\ref{fig:hwp}d),
\begin{equation}
\label{eq:scatter}
\texttt{data}_\ell\bigl[b,\, o_b{+}\delta\bigr] \leftarrow \texttt{present}_\ell\bigl[b,\, t{+}\delta\bigr], \quad \delta \in [0, q),
\end{equation}
for every layer $\ell$. This absolute layout makes retirement free: once a block commits, its KV already resides at the correct locations, subsequent writes simply advance to the new window offset, and committed KV entries are never modified. A per-row EOS immediately deactivates the corresponding sequence.

\paragraph{Discussion.}
Together, GWD and HWP elevate \flowblock{} from a decoding optimization to an execution framework for self-correcting block-wise dLLMs. Unlike training-based approaches such as D2F~\citep{wang2025d2f}, which learn to tolerate incomplete upstream context through distillation, \flowblock{} exploits the same tolerance already present in self-correcting models at inference time. Consequently, it applies to unmodified checkpoints, requires no retraining, and exposes $\thspawn$ as a direct runtime knob for the speed--accuracy trade-off. The additional computation is modest---only $W\!\cdot\!\bl$ extra query positions per forward (e.g.\ $64$ at $W{=}2$, $\bl{=}32$). Combined with HWP's dense batching of asynchronous wavefronts, \flowblock{} is to our knowledge the first engine to make inter-block-parallel dLLM decoding batchable.

\section{Evaluation}
\label{sec:exp}

\subsection{Experimental Setup}
\label{sec:setup}

\paragraph{Implementation.}
We implement \flowblock{} in dInfer, an sgLang-backed inference framework specialized for dLLMs~\citep{inclusionai2025dinfer}. We evaluate with \textbf{LLaDA-2.1-mini}~\citep{inclusionai2026llada21}, an open-weight MoE block-diffusion LLM with T2T self-correction. Unless otherwise stated,  \flowblock{} uses $W{=}2$, retains the model-default decoder thresholds $(\taumask, \tauedit)$, and applies $\thspawn{\approx}0.6$. All experiments run on $8{\times}$GPU (80\,GB) node under data parallelism, with a maximum generation length 2048 and block length 32.

\paragraph{Baselines.} We compare against \textbf{LLaDA-2.0-mini} \citep{inclusionai2025llada2} which uses confidence-threshold parallel unmasking without self-correction, and \textbf{LLaDA-2.1-mini} with its native joint-threshold decoder. Both run in exactly the same engine, KV layout, and evaluation harness as \flowblock{}, eliminating system-level confounds. We also include \textbf{D2F} \citep{wang2025d2f}, a \emph{training-based} inter-block-parallel method, and evaluate its with the same LLaDA-2.0-mini checkpoint---distilled to denoise downstream blocks from incomplete upstream context---under its recommended decoding settings.

\paragraph{Benchmarks and metrics.} We evaluate four math benchmarks: GSM8K \citep{cobbe2021gsm8k}, MATH500 \citep{hendrycks2021math}, Minerva-Algebra \citep{lewkowycz2022minerva}, ASDiv \citep{miao2020asdiv}; and four code benchmarks: HumanEval \citep{chen2021humaneval}, MBPP \citep{austin2021mbpp}, and their EvalPlus variants HumanEval+/MBPP+ \citep{liu2023evalplus}. We report accuracy or pass@1 (\%), \textbf{TPF} (generated tokens per model forward; hardware-independent parallelism), \textbf{TPS} (end-to-end tokens per second), and mean per-request latency (s).

\begin{figure*}[t]
\centering
\includegraphics[width=0.99\textwidth]{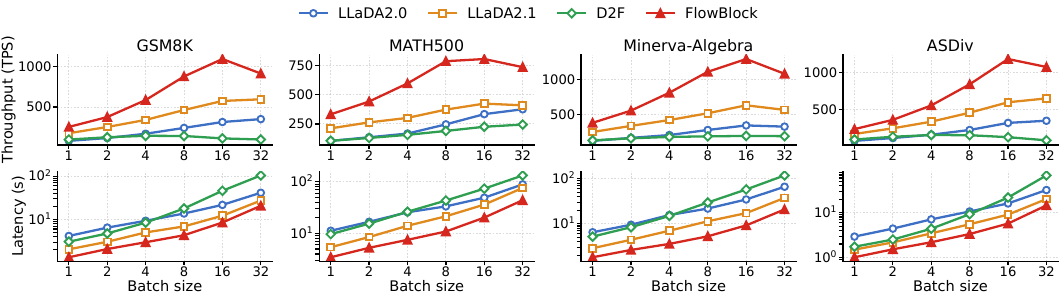}
\caption{Serving throughput and mean per-request latency vs.\ batch size on the four \textbf{math} benchmarks.}
\label{fig:batch-math}
\end{figure*}

\begin{figure*}[t]
\centering
\includegraphics[width=0.99\textwidth]{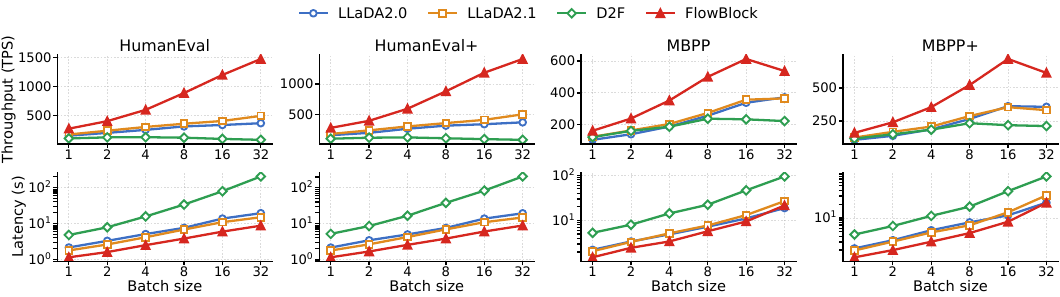}
\caption{Serving throughput and mean per-request latency vs.\ batch size on the four \textbf{code} benchmarks.}
\label{fig:batch-code}
\end{figure*}

\subsection{Main Results}
\label{sec:main-results}

Table~\ref{tab:main} reports batch-size-1 results on all eight benchmarks. Three observations emerge.

\paragraph{Wavefront parallelism compounds with self-correction without sacrificing quality.}

Relative to LLaDA-2.0, LLaDA-2.1's intra-block self-correction raises average TPF from $2.99$ to $4.62$. \flowblock{} extends the same correction mechanism across blocks and reaches $7.23$ TPF, a $1.6\times$ improvement over LLaDA-2.1. End to end, it achieves $2.35\times$ average TPS over LLaDA-2.0 (up to $3.17\times$) and $1.45\times$ over LLaDA-2.1 (up to $1.57\times$), reducing mean latency by $54.5\%$ and $33.3\%$, respectively. These gains do not sacrifice quality: \flowblock{} achieves the best average score ($87.00$, $+1.3$ over LLaDA-2.1), remains within $0.2$ points of the serial baseline on math on average, and improves every code benchmark by $1.3$--$4.9$ points. Because \flowblock{} and LLaDA-2.1 use the same checkpoint, engine, and cache layout, their comparison isolates the scheduling policy.

\paragraph{Training-free overlap outperforms post-training.}
D2F also targets inter-block parallelism, but obtains this ability by distilling a dedicated checkpoint. Its average accuracy is $70.52$, below both serial LLaDA baselines. \flowblock{} improves D2F by $16.5$ average accuracy points and $2.37\times$ average TPS, with up to $3.16\times$ higher TPS and $65\%$ lower latency. This results support exploiting native self-correcting at inference time: \flowblock{} avoids post-training while preserving substantially stronger model quality.


\subsection{Batched Serving}
\label{sec:batch-exp}

Figures~\ref{fig:batch-math}--\ref{fig:batch-code} sweep static batch sizes from $1$ to $32$ on all math and code benchmarks. \flowblock{} leads throughout, and it advantage generally grows with batch size. On Minerva-Algebra, it reaches $1293$ TPS at $B{=}16$ ($2.05\times$ LLaDA-2.1's best). On HumanEval, it reaches $1471$ TPS at $B{=}32$ ($2.95\times$ improvement). Against LLaDA-2.0, the peak gap reaches $4.01\times$ on Minerva-Algebra at $B{=}8$, with per-request latency reduced by up to $77.1\%$. The gain is practically useful: at $B{=}8$, \flowblock{} already matches or exceeds LLaDA-2.1 at $B{=}32$ on GSM8K, MATH500, and Minerva-Algebra, providing comparable throughput with fewer concurrent slots and lower latency. Accuracy remains stable across the sweep.

\paragraph{D2F does not scale effectively with batch size.}
D2F's per-GPU throughput fails to scale with batch size and often declines. On HumanEval, it drops from $113$ TPS at $B{=}1$ to $89$ TPS at $B{=}32$, while mean latency grows to $203$ seconds. \flowblock{} at the same batch reaches a $16.6\times$ throughput advantage and a $95.8\%$ latency reduction. The gap is architectural: D2F parallelizes blocks but does not batch heterogeneous inter-block progress, whereas HWP makes that heterogeneity a first-class execution state.

\begin{figure}[t]
\centering
\includegraphics[width=0.97\columnwidth]{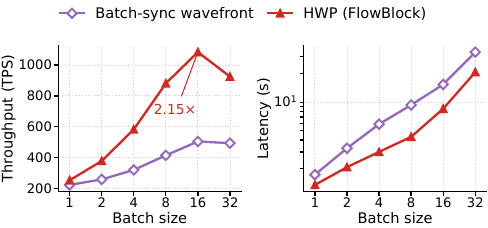}
\caption{HWP vs.\ a batch-synchronous wavefront on GSM8K. Per-sequence wavefronts with dense packing reach $1084$ TPS at $B{=}16$ ($2.15\times$) and reduce latency by $38\%$ at $B{=}32$ (right, log scale); the synchronous variant is throttled by the slowest sequence.}
\label{fig:hwpabl}
\end{figure}

\subsection{Ablation: Heterogeneous Packing}
\label{sec:hwp-exp}

Figure~\ref{fig:hwpabl} isolates HWP by comparing two implementations of the same GWD: a batch-synchronous wavefront with a shared $[L,R)$, \texttt{all()}-gated commit, and $\min$-gated admission; and HWP with per-sequence wavefronts packed into dense forwards. At $B{=}1$ the two are semantically identical. As batch size grows, straggler lockstep throttles the synchronous variant, whereas HWP continues to scale: its TPS advantage is $1.46\times$ at $B{=}2$, $1.82\times$ at $B{=}4$, $2.12\times$ at $B{=}8$, and $2.15\times$ at $B{=}16$. At $B{=}32$, HWP still provides $1.87\times$ higher TPS and $38\%$ lower latency. Accuracy stays within $1.1$ points, confirming that packing changes execution rather than the decoding policy.

\begin{figure}[t]
\centering
\includegraphics[width=0.97\columnwidth]{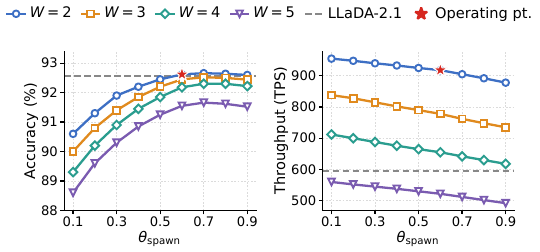}
\caption{Sensitivity to the admission gate $\thspawn$ and window width $W$ on GSM8K at batch size 32. Left: accuracy; right: throughput. Dashed lines mark the LLaDA-2.1 serial reference; the star marks our operating point ($W{=}2$, $\thspawn{=}0.6$).}
\label{fig:threshold}
\end{figure}

\subsection{Sensitivity: the Admission Gate and Window Width}
\label{sec:sens}

Figure~\ref{fig:threshold} sweeps $\thspawn \in [0.1,0.9]$ and $W \in \{2,3,4,5\}$ on GSM8K at batch 32. The two parameters control different trade-offs. The gate primarily affects accuracy: near-ungated admission trails LLaDA-2.1 by $2.0$ points at $W{=}2$ and by up to $4.0$ points at $W{=}5$, whereas moderate gating recovers the serial reference. This confirms the failure mode described in \S\ref{sec:motivation}: drafting from immature upstream context is harmful, and readiness gating suppresses it. Window width primarily determines ccompute cost: throughput decreases monotonically with $W$ because each forward processes $B\!\cdot\!W\!\cdot\!\bl$ query positions. These trends motivate $W{=}2$ and task-specific gate selection. On GSM8K, the smallest gate that clears the serial reference is $\thspawn{=}0.6$, yielding $92.62\%$ accuracy and $918$ TPS, $1.54\times$ over LLaDA-2.1 at serving scale.

\subsection{Robustness to block and generation length.}
Table~\ref{tab:block} varies the block length over $64$, $96$, and $128$ on GSM8K. LLaDA-2.1 degrades in both TPF and accuracy, from $5.17$ to $3.32$ and from $92.57\%$ to $77.41\%$, respectively, as larger blocks require more positions to be resolved from limited intra-block context. D2F is more brittle: its accuracy falls from $53.98\%$ to $22.82\%$, suggesting sensitivity to block geometry. \flowblock{} retains both TPF ($6.93{\to}5.70$, a $1.72\times$ lead over LLaDA-2.1 at $\bl{=}128$) and accuracy ($+2.65$ points at $\bl{=}128$). This supports the intended design: \flowblock{} gains parallelism by overlapping blocks rather than enlarging a block beyond what the model can reliably resolve.

\begin{table}[t]
\centering
\small
\setlength{\tabcolsep}{4.2pt}
\begin{tabular*}{\columnwidth}{@{\extracolsep{\fill}}c l cccc}
\toprule
$\bl$ & Method & TPF & TPS & Lat (s) & Acc (\%) \\
\midrule
\multirow{3}{*}{64}  & LLaDA-2.0 & 1.26 & 54.1 & 28.11 & 32.07 \\
                     & LLaDA-2.1 & 5.17 & 215.0 & 1.83 & 92.57 \\
                     & D2F & 2.57 & 91.4 & 2.97 & 53.98 \\
                     & \textbf{\flowblock{}} & \textbf{6.93} & \textbf{261.2} & \textbf{1.37} & \textbf{92.19} \\
\midrule
\multirow{3}{*}{96}  & LLaDA-2.0 & 1.27 & 54.9 & 33.97 & 22.37 \\
                     & LLaDA-2.1 & 3.98 & 166.3 & 2.27 & 89.99 \\
                     & D2F & 2.29 & 78.5 & 3.30 & 31.54 \\
                     & \textbf{\flowblock{}} & \textbf{6.77} & \textbf{256.8} & \textbf{1.40} & \textbf{91.05} \\
\midrule
\multirow{3}{*}{128} & LLaDA-2.0 & 0.27 & 52.4 & 140.28 & 26.38 \\
                     & LLaDA-2.1 & 3.32 & 140.1 & 2.63 & 77.41 \\
                     & D2F & 1.96 & 63.9 & 3.88 & 22.82 \\
                     & \textbf{\flowblock{}} & \textbf{5.70} & \textbf{216.9} & \textbf{1.68} & \textbf{80.06} \\
\bottomrule
\end{tabular*}
\caption{Block-length scaling on GSM8K ($B{=}1$).}
\label{tab:block}
\end{table}

Table~\ref{tab:length} varies the generation budget from $512$ to $2048$ on GSM8K. \flowblock{} maintains a constant TPF of $6.73$, remains comparable to the LLaDA baselines in accuracy, and preserves its latency advantage at every length ($1.36$\,s versus $2.08$\,s for LLaDA-2.1 and $4.21$\,s for LLaDA-2.0 at length 2048). HWP introduces no generation-length-dependent data-movement overhead because the gather and scatter in Eqs.~\eqref{eq:gather} and~\eqref{eq:scatter} move only the $q$-token active window. D2F remains lower in both quality ($\approx\!85\%$, roughly $7$ points lower) and throughput ($\le\!109$ TPS) across all lengths.

\begin{table}[t]
\centering
\small
\setlength{\tabcolsep}{3.6pt}
\begin{tabular*}{\columnwidth}{@{\extracolsep{\fill}}c l cccc}
\toprule
Length & Method & TPF & TPS & Lat (s) & Acc (\%) \\
\midrule
\multirow{4}{*}{512}  & LLaDA-2.0 & 2.04 & 82.5 & 3.24 & 90.60 \\
                      & LLaDA-2.1 & 4.47 & 176.0 & 1.66 & \textbf{90.98} \\
                      & D2F & 2.96 & 109.1 & 2.34 & 84.00 \\
                      & \textbf{\flowblock{}} & \textbf{6.73} & \textbf{253.3} & \textbf{1.17} & 90.75 \\
\midrule
\multirow{4}{*}{1024} & LLaDA-2.0 & 2.04 & 82.0 & 3.64 & 92.12 \\
                      & LLaDA-2.1 & 4.47 & 177.1 & 1.80 & \textbf{92.57} \\
                      & D2F & 2.82 & 105.7 & 2.60 & 85.29 \\
                      & \textbf{\flowblock{}} & \textbf{6.73} & \textbf{253.8} & \textbf{1.25} & 92.27 \\
\midrule
\multirow{4}{*}{2048} & LLaDA-2.0 & 2.04 & 81.9 & 4.21 & 92.49 \\
                      & LLaDA-2.1 & 4.46 & 174.7 & 2.08 & 92.49 \\
                      & D2F & 2.69 & 99.2 & 3.07 & 85.82 \\
                      & \textbf{\flowblock{}} & \textbf{6.73} & \textbf{249.6} & \textbf{1.36} & \textbf{92.65} \\
\bottomrule
\end{tabular*}
\caption{Generation-length scaling on GSM8K ($B{=}1$).}
\label{tab:length}
\end{table}

\section{Conclusion}
\label{sec:conclusion}

We present \flowblock{}, a training-free framework for wavefront-parallel decoding in self-correcting diffusion language models. Its key insight is that T2T self-correction relaxes inter-block dependencies, turning block finality from a strict prerequisite into a scheduling decision. GWD realizes this insight through readiness-gated admission and a W-shaped block-causal mask that preserves exact frozen-prefix KV reuse. HWP further converts asynchronous per-sequence wavefronts into dense, shape-stable batched computation using per-row gathering, absolute-position KV indexing, and block-diagonal packed attention. Across eight math and code benchmarks, \flowblock{} improves throughput and latency while matching or exceeding baseline accuracy, and scales effectively under batched serving.

\bibliography{aaai2027}

\end{document}